# Dropping Activation Outputs with Localized First-layer Deep Network for Enhancing User Privacy and Data Security

Hao Dong, Chao Wu, Zhen Wei, Yike Guo
Department of Computing, Imperial College London
{hao.dong11, chao.wu, zhen.wei11, y.guo}@imperial.ac.uk

*Abstract*—Deep learning methods can play a crucial role in anomaly detection, prediction, and supporting decision making for applications like personal health-care, pervasive body sensing, etc. However, current architecture of deep networks suffers the privacy issue that users need to give out their data to the model (typically hosted in a server or a cluster on Cloud) for training or prediction. This problem is getting more severe for those sensitive health-care or medical data (e.g fMRI or body sensors measures like EEG signals). In addition to this, there is also a security risk of leaking these data during the data transmission from user to the model (especially when it's through Internet). Targeting at these issues, in this paper we proposed a new architecture for deep network in which users don't reveal their original data to the model. In our method, feed-forward propagation and data encryption are combined into one process: we migrate the first layer of deep network to users' local devices, and apply the activation functions locally, and then use "dropping activation output" method to make the output non-invertible. The resulting approach is able to make model prediction without accessing users' sensitive raw data. Experiment conducted in this paper showed that our approach achieves the desirable privacy protection requirement, and demonstrated several advantages over the traditional approach with encryption / decryption.

## I. Introduction

Deep learning, also known as deep neural networks [1], has been proved to be successful for classification and regression tasks. Once well established, deep learning models exhibited promising performance, which is desired by many real-time prediction / decision-making applications [2], like automatic speech recognition, image recognition, natural language processing, etc.

For these applications, due to hardware limitations (i.e. the computation capability due to the power consumption limitation), it is very expensive to implement deep learning algorithms entirely on users' local devices (sensors, mobiles, and even laptops). As a result, the typical approach is to collect data locally at first and then transmit the data to a remote server / cluster, and then apply the deep networks for training / prediction. However, such approach suffers an important privacy issue, especially for those sensitive data, e.g. patients clinical data (like fMRI images) and sensitive environment data (like video monitoring in public space). While these kind of data are becoming more important for big data analytics (e.g. for personalized health-care), it brings the privacy concern when sending these data directly to a remote model. For example, it would cause severe consequence if a company as a model provider with a lot of users' health-care data like DNA, EHR, fMRI, etc. was invaded by malicious hackers. It also causes another security risk of leaking data during the transmission from user to the model (especially when it's through Internet). It is problematic if we transfer the users' sensitive data directly to a deep learning model hosted on the remote server.

There have been some researches on conducting neural network prediction on encrypted data rather than the original data in the situation so that users / sensors won't share them with remote servers [6], [7]. However, their approach bring in extra computation cost, and it would not be safe once the key for encryption was hacked.

Targeting at this issue, we proposed a novel deep network architecture, in which users don't need to reveal their original data to the model. Instead of the tradition approach shown in Fig 1, we split the layers between local device and the server, and migrate the first layer to local device (as shown in Fig. 2). We apply the activation functions for the first layer locally and then transfer the outputs from the first layer to the server. We then use "dropping activation output" method to make the output non-invertible, which randomly drop some outputs from activation function. In this method, feed-forward propagation and data encryption are combined into one process.

We investigated both invertible and non-invertible activations and their performance in privacy protection. We proved this method can make sure the original data cannot be recovered from the transmitted data. Concretely, the main contributions of this paper are listed as follow:

- We introduced a new deep network architecture that combines neural network and data encryption to solve privacy and data security problem. With privacy preserving feed-forward propagation, server (model provider) can serve for user without directly accessing the user's data.
- We proved and evaluated that dropping few activation outputs can encrypt data for invertible activation function. We also evaluated that ramp function is better than rectifier in term of encryption.
- We proposed privacy preserving error-back propagation, by which server can train a neural network for user without accessing user's data.
- With our method, data compression for data transmission is available when the number of neurons is less than the size of input data.

The paper is organized as follows: Section II firstly introduces the new architecture of deep network and then discusses invertible and non-invertible activation functions, mainly focusing on invertible activation function to encrypt the data. We then give the details of "dropping activation output" method and how it works during feed-forward propagation without any additional computation. Section III provides mathematics proven and explanation for our method. Section IV presents our experiment and its result. We conclude our work in Section V.

## II. METHOD: DROPPING ACTIVATION OUTPUTS WITH LOCALIZED FIRST-LAYER DEEP NETWORK

We firstly introduce our new architecture of localized first-layer deep network, then we introduce the proposed encrypt methods for both invertible activation (like sigmoid) and non-invertible activation (like rectifier). The main content will focus on a method for encrypting data for invertible activation function.

### A. Localized First-layer Deep Network

Fig. 1 illustrates the traditional deep network's architecture, which sends the data from local device to server with encryption / decryption processes, then server uses the original data to compute the result.

The general equation of a single layer can be written as

$$a = f(x * W + b) \quad (1)$$

where $x$ is a row vector of original input data, $W$ is the weight matrix, $b$ is the row vector of biases, $a$ is a row vector of activation outputs and $f$ reflects the activation function such as sigmoid, hyperbolic tangent, softplus, rectifier [9], max-out [18], etc.

From security aspect, the activation outputs $a$ can be captured by the network sniffer, the weight matrix $W$ and bias $b$ can be acquired by hacking the software. As a result, given $a$, $W$, $b$ and $f$, the input data $x$ can be fully reconstructed by Equation (2) [1], where $W^{-1}$ is inverse matrix of $W$ and $f^{-1}$ is the inverse activation function. Therefore, encrypting the activation outputs $a$ is desired for data privacy and security.

$$x = (f^{-1}(a) - b) * W^{-1} \quad (2)$$

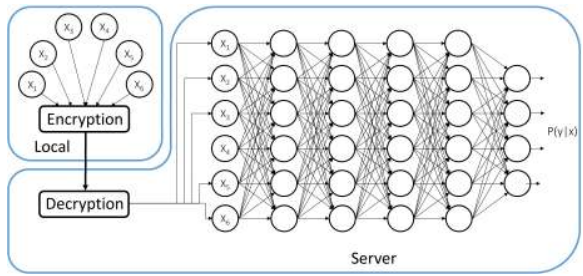

Fig. 1. Implement feed-forward propagation in server; $x$ is the input data, $y$ is the output value, $P(y|x)$ is the probability of $y$ given $x$.

Our new architecture decomposes layers of a neural network between local device and server, as shown in Fig. 2. The local device transfers the activation outputs of first hidden layer to the server, and server can only use the activation outputs to compute the result. We will now discuss how this architecture can enable privacy protection in both non-invertible and invertible activation functions.

[1]If the number of activation outputs equals or greater than the number of input data

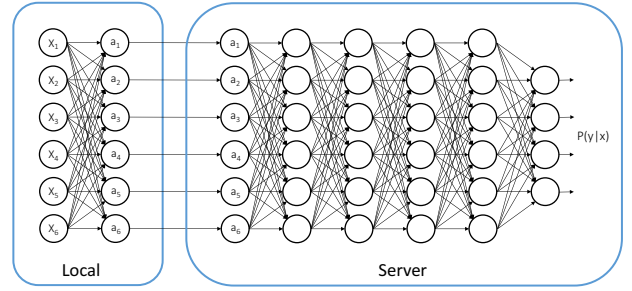

Fig. 2. Spliting neuron network during feed-forward propagation.

### B. Non-invertible Activation Functions

Several up-to-date activation functions are non-invertible [9], [18], for example, rectifier is commonly used in deep neural network, which is able to reach network's best performance without any unsupervised pre-training on unlabeled data [9]. Its output is a linear function of the inputs, so the vanishing gradient problem can be reduced. The rectifier is shown on Equation (3), where $z = x * W + b$. It sets all negative activation outputs to zero.

$$a = f(z) = \begin{cases} 0 & , when\ z < 0 \\ z & , otherwise \end{cases} \quad (3)$$

Non-invertible activation functions naturally can encrypt the activation outputs in some degree. For non-invertible activation functions, the best way to reconstruct input data is to use an approximated inverse activation function $\hat{f}^{-1}$ and the transposed weight matrix $W^T$ (instead of $W^{-1}$), as Equation (4) describes. The mathematical explanation can be found in Section III.

$$\hat{x} = (\hat{f}^{-1}(a) - b) * W^T \quad (4)$$

However, due to the linearity of rectifier, the activation outputs is highly related to the scale of feature patterns from inputs. The general features of original data can be decrypted by Equation (4). To alleviate this problem, we introduce ramp activation function, as Equation (5) shows, where $z = x * W + b$, and $v$ reflects a small value. Smaller $v$ will cut more scale information, so that the activation outputs will contain lesser scale information but more information of features combination (i.e. a set of small feature identifiers). The rectifier and ramp functions will be evaluated in Section IV-A.

$$a = f(z) = \begin{cases} 0 & , when\ z < 0 \\ z & , when\ 0 <= z < v \\ v & , when\ z >= v \end{cases} \quad (5)$$

In the view of information, the non-invertible outputs will result in information loss, so the original data cannot be fully reconstructed. However, from a neural network perspective, the non-invertible function selects features, therefore, information is not lost, and its accuracy can even be improved [9]. Based on this principle, in the rectifying layer even if just one activation output is turned to zero from negative value, the original input data cannot be reconstructed completely. In practice, due to the sparse property of rectifying layer, there will be a large portion of rectifying outputs being zero - therefore, the reconstructed data will be distorted completely.

Hence, data encryption and privacy are realized through non-invertible activation functions, as the input of each hidden layer cannot be reconstructed. Even if all parameters of the model are known, the original inputs still cannot be reconstructed by the server. Hence, privacy is preserved for users. Apart from the rectifier and ramp functions, other activation function such as binary step and max-out [18] are all non-invertible.

## C. Invertible Activation Functions

Different from non-invertible activation functions, reconstruction is solvable when activation function is invertible, e.g. sigmoid, hyperbolic tangent, softplus, etc. The sigmoid and inverse sigmoid is shown in Equation 6 and 7.

$$a = \frac{1}{1 + e^{-(x*W+b)}} \quad (6)$$

$$x = (-ln(\frac{1}{a} - 1) - b) * W^{-1} \quad (7)$$

Therefore, the only ways to encrypt $x$ are modifying $a$, $W$ or $b$. The modification will bring uncertainty to neural network. Therefore, the key is to find a modification method which does not affect the final predicting result.

Due to the sparse behavior of state-of-the-art neural network, accuracy will not be affected by slightly changing activation outputs and weights. First of all, during training, the proposed method would not compromise the learning results, because both Dropout and Dropconnect are the state-of-the-art way to avoid overfitting. For testing / inferencing, non-invertible activation would not affect the result at all. For invertible activation, dropping a few of activation outputs also would not harm the final result, that is because:

1) In theory point of view, Dropout training can be considered as training many sub-networks to do the same job, the result is the averaged of all sub-networks, which is one sort of ensemble learning and recently be proved as a gaussian process [11], [15]. On the other hand, reasonable uncertainty of neural network during testing would not affect the testing result if the network have uncertainty during training [15], [14], [16], [17].

2) In experience point of view, the Dropping probability required by the method (0.5%) is quite low, compared with the Dropping probability during training (usually 20 to 50%), and would not affect the final averaged result of all sub-networks. TABLE I and II show the effect with different dropping probabilities using sigmoid function, as expect, our method would not lead to noticeable performance decreases, as a small dropping probabilities is enough for encryption.

The sparse behavior of neural network can be achieved by the methods of Dropout, Dropconnect, and Autoencoder ([10], [11], [12], [13]). In this paper, we proposed the idea of Dropping activation outputs and Dropping connections. The definitions of Dropping activation outputs and Dropping connections are different from Dropout and Dropconnect. They are applied in the feed-forward propagation, whereas Dropout and Dropconnect are applied in the error-back propagation. A figure of Dropping activation outputs and Dropping connections can be shown in Fig. 3.

***Dropping activation outputs during feed-forward process:*** For encrypting the data during feed-forward propagation process, we purposed a method called Dropping activation output. Mathematically the modified activation output $\hat{a}$ is written as

$$\hat{a} = d \odot f(x * W + b) \quad (8)$$

where, the element-wise multiplication (hadamard product) is denoted by $\odot$, and $d$ is a binary vector randomly, setting some activation outputs to zero.

Setting an activation output to zero is equivalent to removing all connections between a neuron and all input data. We will describe more detail in Section III.

Equation (2) and (4) can be used to reconstruct the data, and we found that similar with non-invertible activation, Equation (4) can better reconstruct the data.

***Dropping connections during feed-forward process:*** Another purposed method is called Dropping connections. In dropping connections, some elements of weight matrix $W$ are set to zero. The modified weight matrix $\hat{W}$ and the modified activation output $\tilde{a}$ are written as

$$\hat{W} = W \odot D \quad (9)$$

$$\tilde{a} = f(x * \hat{W} + b) \quad (10)$$

where, $D$ is a binary matrix with some elements are zero. However, the experiment shows that this method is not effective enough to encrypt the data.

## III. MATHEMATICAL EXPLANATION

If the activation is linear, we have:

$$a = x * W + b \quad (11)$$

$$x = (a - b) * W^{-1} \quad (12)$$

We now discuss why dropping connections won't encrypt as good as dropping activation.

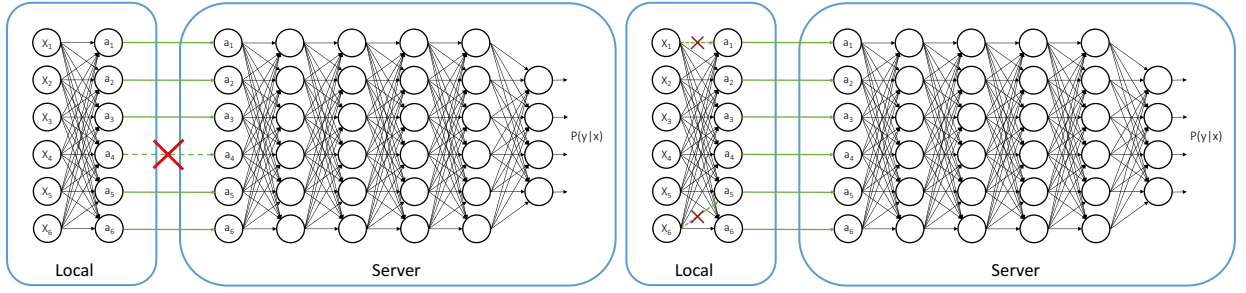

Fig. 3. Dropping activation output (Left) and Dropping connection (Right) during feed-forward propagation.

## A. Dropping connections

Assume $\Delta W$ is the different between $W$ and $\tilde{W}$, i.e. $\Delta W = W - \tilde{W}$, replace $a$ in Equation (12) by Equation (11), let $\tilde{a}$ represent the new $a$ that is calculated through the changed W, then the new $\tilde{x}$ can be written as

$$\begin{aligned}\tilde{x} &= (\tilde{a} - b)W^{-1} \\ &= (x * W \odot D + b - b)W^{-1} \\ &= x(W \odot D)W^{-1} \\ &= x(W - \Delta W)W^{-1} \\ &= x(I - \Delta W W^{-1})\end{aligned} \quad (13)$$

Note in $\Delta W$, only a few entries are non-zero, hence the product of $\Delta W W^{-1}$ is very small, which means it can be neglected. The reason behind it is the product of $WW^{-1}$ is an identity matrix, which means the sum of the product of entries in row W and their corresponding entries in column $W^{-1}$ is 1 when the sum position is diagonal, but the sum is 0 elsewhere. Then $(I - \Delta W W^{-1})$ approximates to an identity matrix. This means the input data is nearly the same as the dropping connection result, written as $\tilde{x} \approx x$, therefore the encrypt performance is poor. Mathematically, it corresponds to the Affine transformation with no change circumstances [28]. This result applies to any $c$-by-$d$ matrix $W$, where $c \leq d$. As $\Delta W_{c \times d}$ is a $c$-by-$d$ matrix with a few entries change, which means only a few terms are affected in the summation of d terms, therefore, $\Delta W_{c \times d} W^{-1}_{c \times d} \approx 0$, hence $I_c - \Delta W_{c \times d} W^{-1}_{c \times d}$ approximate to identity matrix $I_c$, which means $\tilde{x}_{1 \times c} \approx x_{1 \times c}$.

Give an example when W is a 3-by-5 matrix, and $*$ represent entries that are not zero, let $W = \begin{bmatrix} 1 & 3 & 1 & 7 & 2 \\ 2 & 2 & 7 & 5 & 3 \\ 2 & 5 & 3 & 1 & 1 \end{bmatrix}$

- When the changing entry/entries are on the same row in $\Delta W$, then it is the same row that is non-zero in $\Delta W W^{-1}$:

if $\Delta W = \begin{bmatrix} 0 & 3 & 0 & 0 & 0 \\ 0 & 0 & 0 & 0 & 0 \\ 0 & 0 & 0 & 0 & 0 \end{bmatrix}$ or $\begin{bmatrix} 0 & 3 & 0 & 7 & 0 \\ 0 & 0 & 0 & 0 & 0 \\ 0 & 0 & 0 & 0 & 0 \end{bmatrix}$

$$\rightarrow \Delta W W^{-1} = \begin{bmatrix} * & * & * \\ 0 & 0 & 0 \\ 0 & 0 & 0 \end{bmatrix} \quad (14)$$

if $\Delta W = \begin{bmatrix} 0 & 0 & 0 & 0 & 0 \\ 2 & 0 & 0 & 0 & 0 \\ 0 & 0 & 0 & 0 & 0 \end{bmatrix}$ or $\begin{bmatrix} 0 & 0 & 0 & 0 & 0 \\ 0 & 0 & 7 & 5 & 0 \\ 0 & 0 & 0 & 0 & 0 \end{bmatrix}$

$$\rightarrow \Delta W W^{-1} = \begin{bmatrix} 0 & 0 & 0 \\ * & * & * \\ 0 & 0 & 0 \end{bmatrix} \quad (15)$$

if $\Delta W = \begin{bmatrix} 0 & 0 & 0 & 0 & 0 \\ 0 & 0 & 0 & 0 & 0 \\ 0 & 0 & 3 & 0 & 0 \end{bmatrix}$ or $\begin{bmatrix} 0 & 0 & 0 & 0 & 0 \\ 0 & 0 & 0 & 0 & 0 \\ 0 & 5 & 3 & 1 & 1 \end{bmatrix}$

$$\rightarrow \Delta W W^{-1} = \begin{bmatrix} 0 & 0 & 0 \\ 0 & 0 & 0 \\ * & * & * \end{bmatrix} \quad (16)$$

- When changing entries are at different row, then those corresponding rows in $\Delta W W^{-1}$ have non-zero entries.

if $\Delta W = \begin{bmatrix} 0 & 0 & 0 & 0 & 0 \\ 0 & 2 & 0 & 0 & 0 \\ 0 & 0 & 3 & 0 & 0 \end{bmatrix}$

$$\rightarrow \Delta W W^{-1} = \begin{bmatrix} 0 & 0 & 0 \\ * & * & * \\ * & * & * \end{bmatrix} \quad (17)$$

if $\Delta W = \begin{bmatrix} 0 & 0 & 0 & 7 & 0 \\ 0 & 2 & 0 & 0 & 0 \\ 0 & 0 & 3 & 0 & 0 \end{bmatrix} \rightarrow$

$$\rightarrow \Delta W W^{-1} = \begin{bmatrix} * & * & * \\ * & * & * \\ * & * & * \end{bmatrix} \quad (18)$$

$W^{-1}$ is the inverse of W, therefore $WW^{-1} = I_3$ (i.e.

$W_{11}W_{11}^{-1} + W_{12}W_{21}^{-1} + W_{13}W_{31}^{-1} + W_{14}W_{41}^{-1} + W_{15}W_{51}^{-1} = 1$
$W_{11}W_{12}^{-1} + W_{12}W_{22}^{-1} + W_{13}W_{32}^{-1} + W_{14}W_{42}^{-1} + W_{15}W_{52}^{-1} = 0$
$W_{11}W_{13}^{-1} + W_{12}W_{23}^{-1} + W_{13}W_{33}^{-1} + W_{14}W_{43}^{-1} + W_{15}W_{53}^{-1} = 0$
$W_{21}W_{11}^{-1} + W_{22}W_{21}^{-1} + W_{23}W_{31}^{-1} + W_{24}W_{41}^{-1} + W_{25}W_{51}^{-1} = 0$
$W_{21}W_{12}^{-1} + W_{22}W_{22}^{-1} + W_{23}W_{32}^{-1} + W_{24}W_{42}^{-1} + W_{25}W_{52}^{-1} = 1$
$W_{21}W_{13}^{-1} + W_{22}W_{23}^{-1} + W_{23}W_{33}^{-1} + W_{24}W_{43}^{-1} + W_{25}W_{53}^{-1} = 0$
$W_{31}W_{11}^{-1} + W_{32}W_{21}^{-1} + W_{33}W_{31}^{-1} + W_{34}W_{41}^{-1} + W_{35}W_{51}^{-1} = 0$
$W_{31}W_{12}^{-1} + W_{32}W_{22}^{-1} + W_{33}W_{32}^{-1} + W_{34}W_{42}^{-1} + W_{35}W_{52}^{-1} = 0$

$$W_{31}W_{13}^{-1}+W_{32}W_{23}^{-1}+W_{33}W_{33}^{-1}+W_{34}W_{43}^{-1}+W_{35}W_{53}^{-1}=1$$

). However, $\Delta W$ represents change of only a few entries in W, which means only a few terms are affected in the summation of five terms, therefore, $\Delta W W^{-1} \approx 0$, hence $I_3 - \Delta W W^{-1}$ approximate to identity matrix $I_3$, which means $\tilde{x}_{1\times 3} = x_{1\times 3}$.

### B. Dropping activation outputs

In Dropping activation outputs, linear activation is also used in explanation, let the modified activation output be $\hat{a}$ and its corresponding modified input data be $\hat{x}$, then it can be written as

$$\hat{x} = (\hat{a} - b)W^{-1} \qquad (19)$$
$$= (a - \Delta a - b)W^{-1}$$
$$= (a - b)W^{-1} - \Delta a W^{-1}$$
$$= x - \Delta a W^{-1}$$
$$= x(I - \frac{\Delta a W^{-1}}{x})$$

Hence for any activation output vector **a** in size 1-by-$d$ and weight matrix $W$ that has size $c$-by-$d$, where $c \leq d$. If a few entries of **a** change, then the modified vector $\hat{x}$ can be written as a product of 1-by-$c$ original input data x and a matrix $(I - \frac{\Delta a_{1\times d} W_{d\times c}^{-1}}{x_{1\times c}})$, where all diagonal entries are not necessary to be 1. Then $x \not\approx \hat{x}$. Therefore it won't necessary to be no change circumstance in the Affine transformation.

For example, let $a = (a_1, a_2, a_3)$, $\Delta a = (\Delta a_1, \Delta a_2, \Delta a_3)$, and $W_{3\times 2}^{-1} = \begin{bmatrix} w_{11} & w_{12} \\ w_{21} & w_{22} \\ w_{31} & w_{32} \end{bmatrix}$

Then x can be written as:

$$x = (x_1, x_2) = aW_{3\times 2}^{-1} = (a_1, a_2, a_3)\begin{bmatrix} w_{11} & w_{12} \\ w_{21} & w_{22} \\ w_{31} & w_{32} \end{bmatrix} \qquad (20)$$
$$= (a_1 w_{11} + a_2 w_{21} + a_3 w_{31}, a_1 w_{12} + a_2 w_{22} + a_3 w_{32})$$

$\hat{x}$ can be written as:

$$\hat{x} = (\hat{x_1}, \hat{x_2}) = \hat{a}W_{3\times 2}^{-1} \qquad (21)$$
$$= (a_1 - \Delta a_1, a_2 - \Delta a_2, a_3 - \Delta a_3)\begin{bmatrix} w_{11} & w_{12} \\ w_{21} & w_{22} \\ w_{31} & w_{32} \end{bmatrix}$$
$$= (a_1 w_{11} - \Delta a_1 w_{11} + a_2 w_{21} - \Delta a_2 w_{21} + a_3 w_{31} - \Delta a_3 w_{31},$$
$$a_1 w_{12} - \Delta a_1 w_{12} + a_2 w_{22} - \Delta a_2 w_{22} + a_3 w_{32} - \Delta a_3 w_{32})$$

Which means

$$(\hat{x_1}, \hat{x_2})$$
$$= (x_1, x_2)\begin{bmatrix} 1 - \frac{\Delta a_1 w_{11} + \Delta a_2 w_{21} + \Delta a_3 w_{31}}{x_1} & 0 \\ 0 & 1 - \frac{\Delta a_1 w_{12} + \Delta a_2 w_{22} + \Delta a_3 w_{32}}{x_2} \end{bmatrix} \qquad (22)$$

In this case, $\frac{\Delta a_1 w_{11} + \Delta a_2 w_{21} + \Delta a_3 w_{31}}{x_1}$ and $\frac{\Delta a_1 w_{12} + \Delta a_2 w_{22} + \Delta a_3 w_{32}}{x_2}$ are unknown and there is no restriction on them, which means they can be any number, hence it is not an identity matrix.

## IV. EXPERIMENTAL STUDIES

We used MNIST dataset to evaluate our methods, which has a training set of 50k, a validation set of 10k and a test set of 10k. Each image is a 28 x 28 grey-scale digit, with 10 classes in total. We adopted accuracy of classification task to evaluate our model, which is a standard way of evaluating machine learning algorithms. For classification, the definition of accuracy is the percentage of correct prediction on test set.

We further evaluated dropping activation outputs on CI-FAR10 dataset, which is a more challenging dataset compared with MNIST and has 50k training images and 10k test images. Each image is a 32 x 32 RGB image, with 10 classes in total (airplane, automobile, bird, cat, deer, dog, frog, horse, ship, truck).

### A. Non-invertible Activation Functions

**Rectifier** Our experimental network has three rectifying hidden layers, and each layer has 800 neurons, so the model can be represented as (784-800-800-800-10) from input layer to output layer.

Dropout has been applied during training to prevent overfitting. The Dropout probability from the input layer to the first layer is 20%, and the probabilities of other layers are 40%. No weights decay were used. All neural networks were trained by using Adam gradient descent [25] with mini-batches of size 500 and 2000 of epochs. Accuracy of 98.87% was found (under the null hypothesis of the pairwise test with $p = 0.05$).

To evaluate the rectifier, Equation (23) and (24) are used to reconstruct the approximate input data. Fig. 4 shows the original input data and the reconstructed input data from the activation outputs of first hidden layer by using Equation (23). It is clear that the reconstruction was failed.

$$\hat{x} \approx (a - b) * W^{-1} \qquad (23)$$

$$\hat{x} \approx (a - b) * W^T \qquad (24)$$

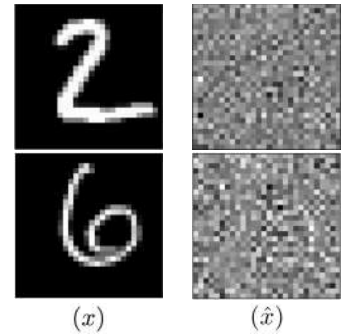

Fig. 4. Reconstructed input data from rectifying output by using $\hat{x} \approx (a-b)*W^{-1}$ (Right); Original input data $x$ (Left); Activation output $a = max(0, x*W+b)$. The KL Divergence between $x$ and $\hat{x}$ for digit 2 and 6 are 1.91 and 1.88.

The input data can be better reconstructed by Equation (24). As the middle column of Fig. 5 demonstrated, even when the reconstructed data were totally distorted, the outline of original input data can still be recognized from the reconstructed data.

Further encryption can be applied by transferring the activation outputs of the second hidden layer, i.e. encrypting the data twice. Right hand side of Fig. 5 shows the reconstructed input data by Equation (24), where the outline of digit cannot be recognized. However, applying more local rectifying layers will lead to higher local computation.

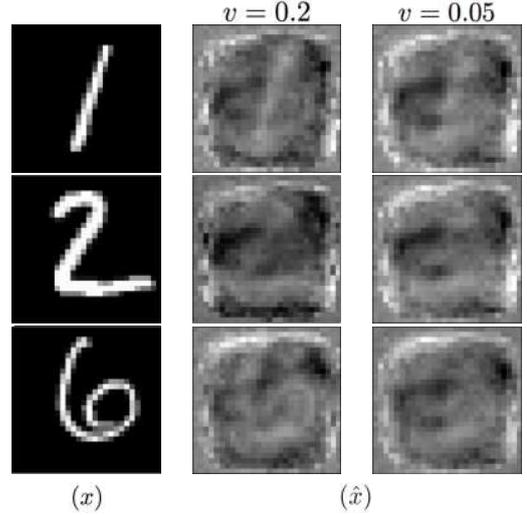

Fig. 6. Reconstructed input data from ramp activation outputs by using $\hat{x} \approx (a-b)*W^T$ (Middle and Right) and Original input data $x$ with $v = 0.2$(Left); The KL Divergence between $x$ and $\hat{x}$ ($v = 0.2$), which are 2.35, 1.54, 1.55 for digit 1, 2, 6, is smaller than that between $x$ and $\hat{x}$ ($v = 0.05$), which are 2.42, 1.63, 1.62.

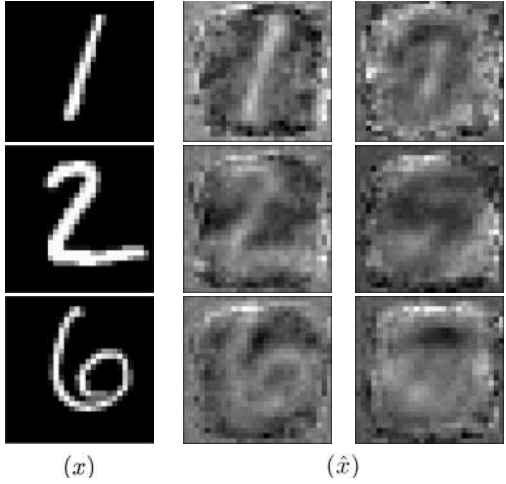

Fig. 5. Reconstructed input data from rectifying outputs by $\hat{x} \approx (a-b)*W^T$: Original input data $x$ (Left); Reconstruct from outputs of first hidden layer (Middle, the KL Divergence between $x$ and $\hat{x}$ (1 st layer) are 2.32, 1.49, 1.51 for digit 1, 2, 6); Reconstruct from outputs of second hidden layer (Right, the KL Divergence between $x$ and $\hat{x}$(2nd layer) are 2.63, 1.58, 1.56 which are higher than $\hat{x}$ (1st layer)).

**Ramp** To let the activation outputs of first hidden layer contain lesser scale information, ramp activation function can be applied. The two testing networks are the same with the previous rectifier network, except for the activation function of first layer, which are ramp with $v$ of 0.2 and 0.05 respectively. The accuracies of these two networks are both 98.87%, which are approximately the same as the previous pure rectifier neural network.

According to Fig. 6, smaller $v$ can better encrypt the input data, and the topology of input data cannot be recognized from the reconstructed data.

The above experiments can be summarized as below:

- With the purposed architecture, no matter what kind of activation function is used, without knowing the model parameters in local device, the original input data cannot be reconstructed from the activation output.
- Even knowing the model parameters in local device, a single rectifying layer is able to encrypt the input data. Better encryption can be done by applying more rectifying layers before transferring the activation outputs.
- Further encryption to the data can be done through Ramp activation function as it reduces the scale of feature pattern.

### B. Invertible Activation Functions - Dropping Activation Outputs

For invertible activation function, the first experiment is about Dropping activation outputs. We consider a networks of 3 hidden layers of 800 units each, the activation function is sigmoid, which is invertible. All hyper-parameters and training methods are set to be the same as the previous rectifier neural network.

As a few of activation outputs are set to zero, the inverse sigmoid at Equation (7) becomes insolvable, since a number can not be divided by zero. An approximate reconstruction can be achieved by Equation (25) and (4), i.e. when $a$ is out of range, the approximate inverse number is set to zero [1].

$$f^{-1}(a) = \begin{cases} 0 & , when \ a \geq 0, a \leq 0 \\ -ln(\frac{1}{a} - 1) & , otherwise \end{cases} \quad (25)$$

Fig. 7 shows the reconstructed input data under dropping probabilities $p$ of 0.5%, 1% and 2%. It is clear that the reconstructed input data were distorted, and no further distortion was found as the dropping probability increase.

Therefore, dropping a few activation outputs is enough to encrypt the input data, and does not lead to noticeable performance decreases as Table. I shows. In addition, adding some noise to a few activation outputs can also lead to the same result. If we choose Equation (2) for reconstruction, the reconstructed input data will be show in Fig. 8, which is worse than Fig. 7.

For CIFAR10, we used a convolutional neural network (CNN) for this task. The network architecture is as follows:

[1]Expect zero, the following constants have been explored: $\{-10, -1, -0.5, 0, 0.5, 1, 10\}$, we found them all giving similar results as Fig 7 shows.

## TABLE I
RANDOMLY SETTING ACTIVATION OUTPUTS TO ZERO IN FIRST HIDDEN LAYER DURING FEED-FORWARD PROPAGATION OVER 100 TRAILS

| $p$ | Accuracy(%) | Standard Derivation | Max/Min Accuracy (%) |
|---|---|---|---|
| 0% | **98.7900** | NA | NA |
| 0.5% | **98.8026** | 0.000221 | 98.86 / 98.74 |
| 1% | **98.8081** | 0.000294 | 98.88 / 98.73 |
| 2% | **98.8093** | 0.000356 | 98.92 / 98.73 |
| 3% | **98.7978** | 0.000373 | 98.89 / 98.69 |
| 5% | 98.7721 | 0.000419 | 98.87 / 98.65 |
| 10% | 98.6991 | 0.000500 | 98.87 / 98.57 |

## TABLE II
RANDOMLY SETTING ACTIVATION OUTPUTS TO ZERO IN FIRST CONVOLUTIONAL LAYER DURING FEED-FORWARD PROPAGATION OVER 100 TRAILS

| $p$ | Accuracy(%) | Standard Derivation | Max/Min Accuracy (%) |
|---|---|---|---|
| 0% | **84.5640** | NA | NA |
| 0.1% | **84.5641** | 0.000409 | 84.68 / 84.51 |
| 0.5% | 84.5535 | 0.000635 | 84.78 / 84.48 |
| 1% | 84.5313 | 0.000787 | 84.70 / 84.35 |
| 2% | 84.4696 | 0.001037 | 84.71 / 84.16 |
| 3% | 84.4180 | 0.000859 | 84.23 / 84.75 |
| 5% | 84.3613 | 0.001230 | 84.68 / 84.09 |

N64F5S1> Sigmoid> F3S2> LRN> N64F5S1> ReLU> LRN> F3S2> D384> ReLU> D192> ReLU> D10> Softmax, where N64F5S1 represents a CNN with 64 filters of size 5 and of stride 1, F3S2 represents Maxpooling with size of 3, stride of 2, and D10 represents fully-connective layer with unit of 10, LRN represents local response normalization. The model is trained by 50000 epochs with learning rate of 0.0001 and Adam gradient descent. We randomly dropped activation outputs on the 1st convolutional layer. The result can be seen in Table. II.

The experiments can be summarized as follow:

- For invertible activation function, dropping a few activation outputs during feed-forward propagation can provide data encryption and privacy.
- To make the function insolvable, the activation value needs to be set out of its reasonable range, for example, setting the value out of $(0, 1)$ for sigmoid, and $(-1, 1)$ for hyperbolic tangent.
- Instead of setting activation outputs out of its reasonable range, adding small noise values to few activation output can have similar impact.

### C. Invertible Activation Functions - Dropping Connections

Randomly setting a part of weight values to zero can indirectly modify the activation outputs. However, according to our experiment by using MNIST dataset, when combining sigmoid function and dropping activation outputs during feed-forward process, encryption does not appear and the reconstructed data is almost the same with the original input data.

### D. Autoencoder

Small number of activation output can reduce both computation and communication cost for local device. In that case, Autoencoder can be applied.

Even for invertible activation function, when the number of activation outputs is smaller than the number of input data, the original input data cannot be reconstructed correctly from its activation outputs, as it uses smaller dimension to represent the original data. The input data can be approximately reconstructed by Equation (7).

Fig. 9 shows the input data which is reconstructed from Autoencoder with sigmoid activation function. In this case, the number of the neuron is half of the number of input data. In other words, half of the information of original input data can be reconstructed, noting the shape of digit image can still be identified clearly. Therefore, an Autoencoder with non-invertible activation function or dropping activation output are recommended, so as to provide both data encryption and data compression.

### E. Privacy preserving

Here we analyze the privacy preserving property of our method, by discussing how it works in bruteforce attack and honest-but-curious model.

Assume we drop $M$ neurons, as we discussed before, the computation raised exponentially, the attackers have to pre-compute possible activation results. Take sigmoid function as an example, as the outputs of neurons are continuous values, assume attacker pre-define $N$ possible output values, and the algorithm drop $M$ neurons of the outputs, the number of possible combinations is $N^M$. In our sigmoid experiment when dropping probability is 0.5%, $M$ is 4, if define $N$ to 100, the number of possible combinations is 100 millions, after reconstructing 100 millions images, the attacker also need an algorithm to recognize the good image. If we used rectifier, the attacker would not know where and how many neurons are dropped by our method, which mean the number of possible combinations will become significantly larger.

We illustrate this by conducting a bruteforce experiment on MNIST classifier with sigmoid function. There are 800 outputs, $M$ is 4 (0.5%). We separated [0, 1] to 101 gaps, i.e. from 0, 0.01, 0.02 to 1. Therefore, there is 104060401 possible inputs, it takes 5083 hours to compute all possibilities on a Titan X Pascal GPU. Note that, the attacker needs an extra algorithm to select one data from 104060401 data, but as the attacker do not know the content of data, it is difficult to define the criteria.

In honest-but-curious model, for invertible activation function, the intact activation outputs can be obtained if local device inputs the same data into the network for multiple times, and the dropping positions do not overlap. The probability of the dropping position do not overlap on second time with the first time is $C_{N-M}^M / C_N^M$. As $M$ is relatively smaller than $N$, so the original input data has high probability to be decrypted if the data is input to the network twice.

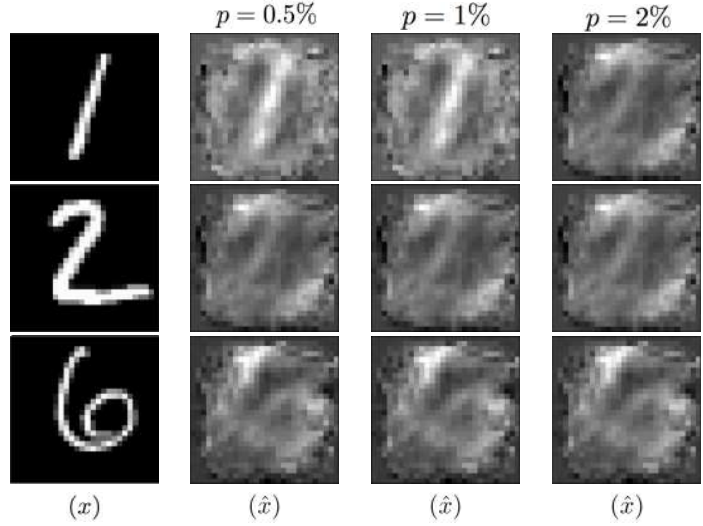

Fig. 7. Equation (25) and (4) are used to reconstruct input data from first sigmoid hidden layer after dropping activation outputs. The first column is the original input data $x$; the dropping probabilities increase from 0.5% to 2% from second to forth column. Note that 0.5% is only 4 neurons in the layer of 800 neurons. The KL Divergence between $x$ and $\hat{x}(p=0.5\%)$ are 2.03, 1.34, 1.35 which are similar with $\hat{x}(p=1\%)$ (2.02, 1.35, 1.35) and $\hat{x}(p=2\%)$ (2.04, 1.35, 1.36).

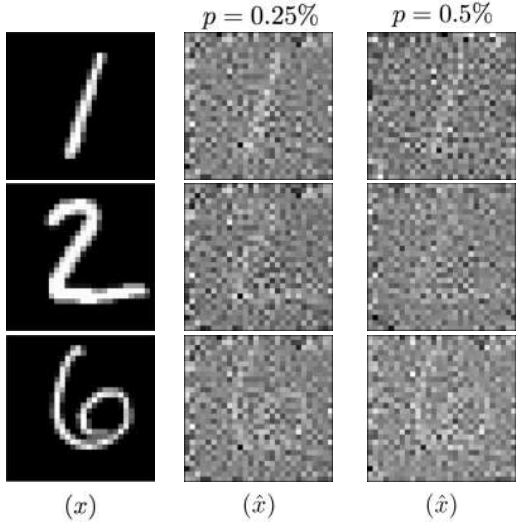

Fig. 8. Reconstructed input data from first sigmoid hidden layer after drop activation outputs by using Equation (25) and (2). Original input data $x$ (The first column); From second to third columns, the dropping probabilities increase from 0.25% to 0.5%. The KL Divergence between $x$ and $\hat{x}(p=0.25\%)$ are 2.34, 1.51, 1.45, and $\hat{x}(p=0.5\%)$ are 2.49, 1.52, 1.46, which are higher than Fig.7

As a result, in practice, to prevent such situation, we drop the same position for the same data. We used the max value in the data as the random seed to select the dropping position.

We need to mention that, for non-invertible activation function, as the activation outputs are dropped inherently, even without using dropping activation outputs, the honest-but-curious model can not work.

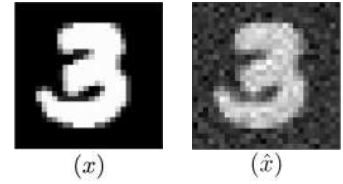

Fig. 9. Reconstructed input data from sigmoid outputs by using Equation (7) (Right); Original input data $x$ (Left). The number of neuron is the half of the number of input data.

*F. Speed*

To verify the impact of implementing Sigmoid function on user local devices, we did a system test to measure it. We used Metal framework [1] to develop the Sigmoid function on GPU (in Swift language):

```
kernel void sigmoid(const device float
  *inVector [[ buffer(0) ]],
  device float *outVector [[ buffer(1) ]],
  uint id [[ thread_position_in_grid ]]) {
    outVector[id] = 1.0 /
    (1.0 + exp(-inVector[id]));
}
```

The device we used is iPhone 6 with iOS 10. $NSDate()$ was used to measured the time lapse of computing the first layer activations. The input we used is a vector $x$ where $|x| = 40000$ (200x200 pixels grayscale images), and we applied $Sigmoid(W\dot{x} + b)$ with $W$ (size 40000x1000) and $b$ (size 1000) for 1000 neurons. Below is a sample output of time lapses measured, **1300ms** is calculated to be the average latency induced.

[1] https://developer.apple.com/metal/

```
D/Sigmoid(601): onActivation: begin
D/Sigmoid(601): onActivation: 1311 ms
D/Sigmoid(601): onActivation: end, 1311 ms
```

## V. CONCLUSION

In this paper, we proposed a new architecture for deep network, with the localized first layer of the network. We investigate how this architecture can support better privacy protection in model prediction. Invertible activation function through Dropping activation outputs during feed-forward propagation are proved to be able to encrypt the original input data and preserving privacy. The whole encryption process can be improved through combining feed-forward propagation and data encryption into one process, which means no need for a specialized data encryption process on the local device, nor data decryption process on the server.

During the encryption process, both invertible and non-invertible activation functions have been discussed and mathematically proved possible to do encryption. In error-back propagation, splitting the neural network into local device and server can provide data privacy during training. In other words, the server is able to provide model learning service by using error-back propagation, without accessing the original input data from the local device.

## ACKNOWLEDGMENT

The authors would like to thank Charles R. Johnson from William and Mary College and Xinman Ye from University of Cambridge, for their helpful comments and suggestions on the manuscript. Hao Dong is supported by the OPTIMISE Portal.